\documentclass[12pt,a4paper]{article}

\usepackage[margin=0.8in]{geometry}
\usepackage{setspace} 
\usepackage{authblk}
\usepackage{times}
\usepackage{cite}
\usepackage{graphicx}
\usepackage{subfig}
\usepackage{upgreek}
\usepackage{amsthm,amsmath,amssymb,bm}

\usepackage{multirow}
\usepackage{tcolorbox}
\usepackage{lettrine}
\usepackage{color}
\usepackage[hidelinks,colorlinks=false]{hyperref}
\usepackage[ruled,linesnumbered]{algorithm2e}

\newtheorem{definition}{Definition}

\graphicspath{{figures/}}

\begin{document}

\title{{\huge\textbf{Mean-Shift Theory and Its Applications in Swarm Robotics}}\\
{A New Way to Enhance the Efficiency of Multi-Robot Collaboration}} 

\author{
	Guibin~Sun, 
	Jinhu L\"{u}, \emph{Fellow, IEEE}, 
	Kexin~Liu, 
    Zhenqian Wang, 
	and Guanrong Chen, \emph{Life Fellow, IEEE}
}

\date{}
\maketitle

\begin{figure}[!h]
    \centering
    \includegraphics[width=\linewidth]{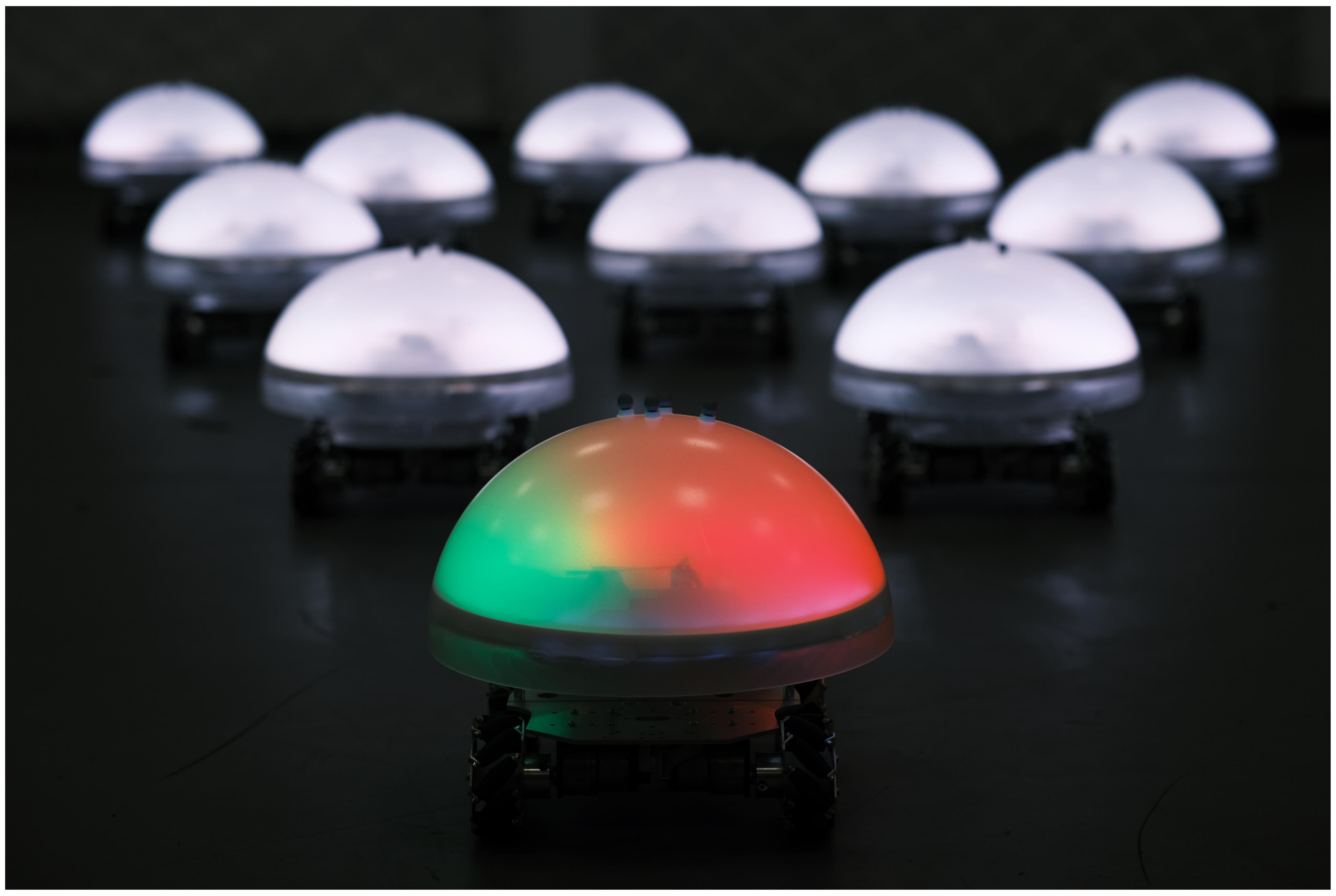}
\end{figure}

\newpage


\lettrine[lines=2]{S}{warms} evolving from collective behaviors among multiple individuals are commonly seen in nature, which enables biological systems to exhibit more efficient and robust collaboration. 
Creating similar swarm intelligence in engineered robots poses challenges to the design of collaborative algorithms that can be programmed at large scales. 
The assignment-based method has played an eminent role for a very long time in solving collaboration problems of robot swarms. 
However, it faces fundamental limitations in terms of efficiency and robustness due to its unscalability to swarm variants. 
This article presents a tutorial review on recent advances in assignment-free collaboration of robot swarms, focusing on the problem of shape formation. 
A key theoretical component is the recently developed \emph{mean-shift exploration} strategy, which improves the collaboration efficiency of large-scale swarms by dozens of times. 
Further, the efficiency improvement is more significant as the swarm scale increases. 
Finally, this article discusses three important applications of the mean-shift exploration strategy, including precise shape formation, area coverage formation, and maneuvering formation, as well as their corresponding industrial scenarios in smart warehousing, area exploration, and cargo transportation. 

\section*{Introduction}

Robots constitute a prominent part of industrial automation systems, and they are universally recognized as synonymous with the industrial revolution. 
In recent years, collaborative robotics has received considerable attention due to its tremendous potential across many application domains \cite{Wang2019CSM,Grau2021IEM,Sun2023NC}, such as cargo transportation \cite{Alonso2017IJRR}, environmental monitoring \cite{Shah2020SCIROB}, targeted delivery \cite{Chen2025SCIROB}, and aerial manipulation \cite{Cao2025NATURE}. 
Researchers are inspired by natural swarm intelligence, such as insect swarms and bird flocks \cite{Dorigo2021IEEEPROC,Ling2019NEE,Duan2023NSR}, to empower robot swarms with efficient and robust capabilities. 
As a result, collaborative robots become more flexible and fault-tolerant than a single robot, and can better adapt their behavior to changes in tasks and environments. 

Shape formation is one fundamental task for robot swarms, which enables robots to execute overwhelming tasks in a well-organized manner. 
The purpose of shape formation is to direct a group of robots starting from any initial configuration to form a desired geometric shape by using local information from neighbors \cite{Zhao2019CSM}. 
As a remarkable example, product sorting problems in smart warehousing can be mapped to shape formations \cite{Wang2020TRO}. 
It is reported that a swarm of 350 robots can sort 18,000 express packages per hour, saving 70\% of labor \cite{LibiaoRobot}. 

\begin{table}[!b] 
    \centering
	\caption{Summary of typical shape formation methods.} 
	\label{Tab_summary} 	
	\begin{tabular}{cccccccc}
		\hline
        \multirow{2}{*}{\textbf{Method}} & \multirow{2}{*}{\textbf{Category}} & \textbf{Unique} & \textbf{Topology} & \textbf{Dead-} & \textbf{Forming} & \textbf{Swarm} & \multirow{2}{*}{\textbf{Reference}} \\
         & & \textbf{identity} & \textbf{constraint} & \textbf{lock} & \textbf{shape} & \textbf{scale} &  \\
		\hline
        \multirow{4}{*}{\textbf{graph-based}} & position & \checkmark & \checkmark & - & simple &  & \cite{Coogan2012AUT,Nguyen2020TRO} \\
        \cline{2-6} \cline{8-8}
         & displacement & \checkmark & \checkmark & - & simple & typically & \cite{Fax2004TAC,Dehghani2016AST} \\
        \cline{2-6} \cline{8-8}
         & distance & \checkmark & \checkmark & - & simple & $\le$ 10 & \cite{Krick2009IJC,Bae2021TAC} \\
        \cline{2-6} \cline{8-8}
         & bearing & \checkmark & \checkmark & - & simple &  & \cite{Zhao2015TAC,Li2022TC} \\
        \hline
         & centralized & \multirow{2}{*}{\checkmark} & \multirow{2}{*}{-} & \multirow{2}{*}{-} & \multirow{2}{*}{complex} & typically & \multirow{2}{*}{\cite{Alonso2012IJRR,Yu2013CDC,Wang2016TNNLS}} \\
        \textbf{assignment} & assignment &  &  &  &  & $\le$ 100 &  \\
        \cline{2-8}
        \textbf{-based} & distributed & \multirow{2}{*}{\checkmark} & \multirow{2}{*}{-} & \multirow{2}{*}{\checkmark} & \multirow{2}{*}{complex} & \multirow{2}{*}{large} & \multirow{2}{*}{\cite{Morgan2016IJRR,Hu2020TRO,Chu2023TASE,Li2024TIE}} \\ 
         & assignment &  &  &  &  &  &  \\
        \hline
         & edge- & \multirow{2}{*}{\checkmark} & \multirow{2}{*}{-} & \multirow{2}{*}{-} & \multirow{2}{*}{complex} & \multirow{8}{*}{large} & \multirow{2}{*}{\cite{Rubenstein2014Science}} \\
         & following &  &  &  &  &  &  \\
         \cline{2-6} \cline{8-8}
         & reaction- & \multirow{2}{*}{-} & \multirow{2}{*}{-} & \multirow{2}{*}{-} & \multirow{2}{*}{unstable} &  & \multirow{2}{*}{\cite{Slavkov2018SR,Gardi2022NC}} \\
        \textbf{behavior-} & diffusion &  &  &  &  &  &  \\
        \cline{2-6} \cline{8-8}
        \textbf{based} & random- & \multirow{2}{*}{-} & \multirow{2}{*}{-} & \multirow{2}{*}{-} & \multirow{2}{*}{profile} &  & \multirow{2}{*}{\cite{Alhafnawi2021RAL}} \\
         & walk  &  &  &  &  &  &  \\
         \cline{2-6} \cline{8-8}
         & artificial- & \multirow{2}{*}{-} & \multirow{2}{*}{-} & \multirow{2}{*}{\checkmark} & \multirow{2}{*}{simple} &  & \multirow{2}{*}{\cite{Hsieh2008ROBOTICA,Hou2012IET,Vickery2021ICARA}} \\
         & potential  &  &  &  &  &  &  \\
		\hline
	\end{tabular}
\end{table}

In shape formation, the way of describing a desired shape is an important factor in determining the collaboration strategy. 
One class of strategies widely studied is to preset the shape constraints by such as inter-robot position \cite{Coogan2012AUT,Nguyen2020TRO}, displacement \cite{Fax2004TAC,Dehghani2016AST}, distance \cite{Krick2009IJC,Bae2021TAC}, and bearing \cite{Zhao2015TAC,Li2022TC} (Table~\ref{Tab_summary}). 
These relative constraints among cooperative robots are often described by a graph, so such strategies are called graph-based formation. 
Once the desired constraints are determined, robots can be controlled through the feedback of the errors between their predefined and current states \cite{Sakurama2020Aut}. 
Conventional feedback-based methods require each robot to be uniquely labeled to match a predefined topology. 
Specifically, the robot is assigned a unique identity to recognize whom the topological neighbors are. 
However, this requirement lacks robustness against joining or failing robots \cite{Chen2015TC,Queralta2019PCS}. 
Moreover, robots may be unable to identify the labels of their neighbors in motion, especially if wireless communication is not available \cite{Sakurama2020Aut}. 

Another class of methodologies for shape formation widely adopts a procedure of goal assignment in either centralized \cite{Alonso2012IJRR,Yu2013CDC,Wang2016TNNLS} or distributed fashion \cite{Morgan2016IJRR,Hu2020TRO,Chu2023TASE,Li2024TIE}  (Table~\ref{Tab_summary}). 
In these methods, the task of shape formation can be divided into two subtasks, i.e., assigning goal positions to each robot and planning a collision-free path to that goal \cite{Wang2020TRO}. 
Although widely adopted, assignment-based methods suffer from some fundamental limitations. 
The first is the tradeoff between computational complexity and motion efficiency. 
Centralized assignment has high motion efficiency because, once a robot is assigned a unique goal, it can move efficiently to the target while avoiding collision with other robots \cite{Mora2011ICRA}. 
Unfortunately, the computational complexity increases rapidly as the swarm scale increases, making it inefficient to support large-scale swarms \cite{Morgan2016IJRR}. 
In comparison, distributed assignment can support large-scale robot swarms by decomposing the centralized assignment into multiple lower-scale ones. 
However, conflicts among local assignments are inevitable due to the use of local information, which must be resolved by sophisticated algorithms such as task swapping \cite{Wang2020TRO}. 
 
The other important type of shape formation is based on local behavior (Table~\ref{Tab_summary}). 
A representative method that can form complex shapes by using thousands of robots is the edge-following strategy \cite{Rubenstein2014Science}. 
This strategy, however, has low efficiency since only the robots on the edge of the swarm can maneuver while those inside the swarm must standstill. 
In contrast, the method inspired by reaction-diffusion morphogenesis can form emergent shapes spontaneously \cite{Slavkov2018SR,Gardi2022NC}. 
However, the shape formed in emergent morphogenesis cannot be predefined by users \cite{Alhafnawi2021RAL}. 
Recent research based on local behaviors can achieve aggregation of swarm robots on the profile of the shape \cite{Alhafnawi2021RAL}. 
Since robots cannot self-localize relative to the shape, they need to use a random-walk strategy to search for shape edges, which usually leads to low motion efficiency \cite{Sun2023NC}. 
Another popular behavior-based method is based on artificial potential \cite{Hsieh2008ROBOTICA,Hou2012IET,Vickery2021ICARA}, where a robot is driven to enter the shape by an attractive force and avoid collision by a repulsive force. 
These attraction-repulsion methods, however, may easily get stuck at local minima, making it difficult to form complex shapes. 

The purpose of this article is to provide a tutorial review of our recent research \cite{Sun2023NC,Zhang2024RAL} on mean-shift based shape formation of homogeneous robot swarms to industrial practitioners. 
This strategy does not require goal assignments but instead is based on the idea of mean-shift exploration. Specifically, when a robot predicts competition with neighbors for the same goal positions, it will selflessly give up its current position by exploring the highest density of unoccupied positions in the shape. 
This idea is realized by an adaptive version of the mean-shift algorithm \cite{Fukunaga1975TIT,Cheng1995PAMI}, which is an optimization technique widely used in machine learning for searching the maxima of a density function. 

\begin{figure}[!t]
    \centering
    \includegraphics[width=\linewidth]{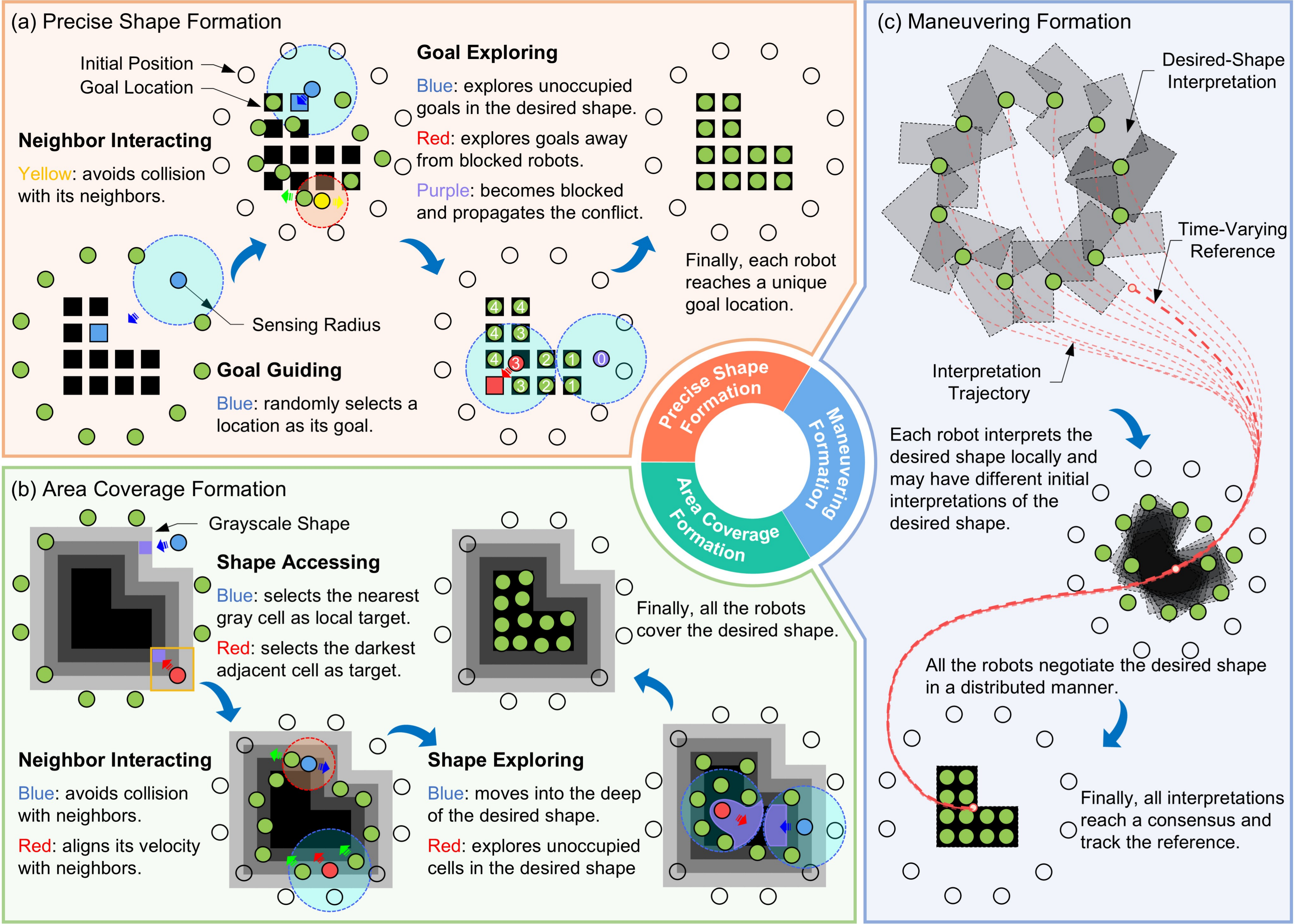}
    \caption{An overview of mean-shift collaboration. 
    (a)~Precise shape formation.  
    (b)~Area coverage formation. 
    (c)~Maneuvering formation.} 
    \label{Fig_overview}
\end{figure}

This article addresses three essential applications of the mean-shift theory in the field of shape formation of homogeneous robot swarms, summarized as follows. 
\begin{itemize}
    \item \emph{Precise shape formation}: Consider a desired shape consisting of a set of distinct goal positions that determine the desired positions of robots in the shape. 
    The control objective is to move a group of robots from their initial positions to the goal positions in the shape through local interactions (Figure~\ref{Fig_overview}(a)). 
    Here, every robot corresponds to a unique goal position in the desired shape. 
    This task is essential for robot swarms to achieve complicated applications such as smart warehousing (Figure~\ref{Fig_precise}(b)). 
    \item \emph{Area coverage formation}: Consider a formation shape specified by a binary grid, in which the black area is where robots are expected to move to. 
    The control objective is to steer a robot swarm starting from some initial configuration to cover a user-specified shape area via local interactions (Figure~\ref{Fig_overview}(b)). 
    Different from precise formations, area coverage formation requires robots to cover a shape area, rather than moving to specific positions. 
    This task exhibits strong robustness to individual faults and has promising potential in area exploration (Figure~\ref{Fig_coverage}(d)). 
    \item \emph{Maneuvering formation}: 
    The objective of maneuvering formation is to drive a group of robots to collectively maneuver so that the centroid and orientation of the swarm can be changed while maintaining a desired shape (Figure~\ref{Fig_overview}(c)). 
    This task requires introducing a small number of informed robots who know a user-specified trajectory of the shape. 
    This finds important applications in achieving complex tasks such as cargo transportation (Figure~\ref{Fig_maneuver}(c)). 
\end{itemize}

\begin{tcolorbox}[colback=black!8!white,colframe=black!8!white,top=0.5em,bottom=0.5em]    
    \subsection*{Notation for Swarm Networks} 
    Consider a group of $n_{\rm robot}$ robots in $\mathbb{R}^2$.  
    Each robot is treated as a circle and is modeled by a single integrator $\dot p_i=v_i$. 
    Here, $p_i\in \mathbb{R}^2$ is the position of robot $i$, and $v_i$ is the velocity command to be designed and limited by a threshold $v_{\rm max}$. 
    Note that $v_i$ is easily converted into linear and angular speeds for nonholonomic robots by \cite{Zhao2018TAC}. 
    Two robots could share information with each other if the distance between them is less than $r_{\rm sense}$. 
    The interaction network among robots is described by an undirected graph $\mathcal{G}=(\mathcal{V},\mathcal{E})$, which is composed of a vertex set $\mathcal{V} = \{1,\ldots,n_{\rm robot}\}$ and an edge set $\mathcal{E} = \{(i,j): \|p_i-p_j\| < r_{\rm sense}, j \neq i\}$. 
    The set of neighbors for robot $i$ is denoted as $\mathcal{N }_i=\{j \in \mathcal{V}: (i,j) \in \mathcal{E}\}$. 
\end{tcolorbox}

The notation for swarm networks used in this article is given in ``Notation for Swarm Networks''. 

\section*{Mean-Shift Theory}

\begin{figure}[!b]
    \centering
    \includegraphics[width=\linewidth]{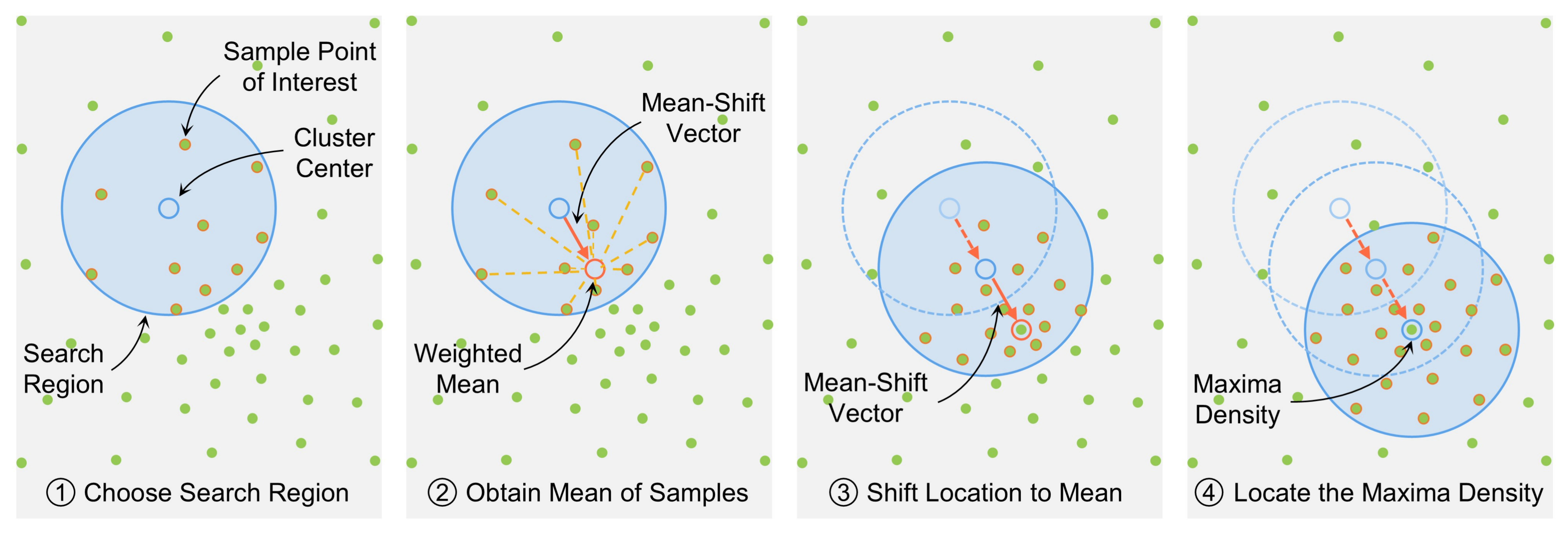}
    \caption{An illustration of the mean-shift process. }
    \label{Fig_meanshift}
\end{figure}

Mean shift is a nonparametric and iterative process for estimating the gradient of a density function. 
Equivalently stated, mean shift is an optimization technique for locating the maxima of a density function represented by a set of samples. 
To illustrate this, an example is shown in Figure~\ref{Fig_meanshift}, describing the iterative process for locating the maximum density. 
In each iteration, the mean-shift vector shifts the cluster center to the weighted mean of the sample points in its neighborhood until the density maxima are located. 

Mean-shift theory was first introduced by Fukunaga and Hostetler \cite{Fukunaga1975TIT} in 1975. 
Later, Cheng \cite{Cheng1995PAMI} advanced the theory by incorporating the kernel function, further enhancing its applicability. 
Mean-shift technique is widely used in computer vision, such as image segmentation and object tracking \cite{Zhang2025NN}. 
More recently, this technique has been successfully adapted for multi-robot collaboration. 
A representative study is the recently developed mean-shift exploration strategy for multi-robot coverage formation \cite{Sun2023NC}, which is subsequently extended to precise formation \cite{Zhang2024RAL} and multi-shape formation control \cite{Li2025TASE}. 
The precise definition and useful properties are given as follows. 

\subsection*{\emph{Mean-Shift Definition}}

This review is based on the generalized version of the mean shift theory introduced by Cheng \cite{Cheng1995PAMI}. 
There are two fundamental concepts in the mean shift, which are precisely defined in ``Key Definitions in Mean-Shift Theory''. 
Here, we employ the definition in terms of a profile (Definition~\ref{Def_profile}), which can improve the clarity of our analysis without departing from the original definition. 
In this way, we can define the sample mean with a profile (Definition~\ref{Def_meanshift}). 
Now, we are ready to present the mean-shift definition. 

\begin{tcolorbox}[colback=black!8!white,colframe=black!8!white,top=0.5em,bottom=0.5em]    
    \subsection*{Key Definitions in Mean-Shift Theory}
    \begin{definition}[\bf Profile] \label{Def_profile}
        A function $k:\mathbb{R}^+\rightarrow \mathbb{R}^+$ is said to be a profile if it is: 
        1) nonnegative, 2) monotonically nonincreasing, i.e., $k(a)>k(b)$, if $a<b$, 3) piecewise continuous, and  4) $\int_0^\infty k(r){\rm d}r < \infty$. 
    \end{definition}

    \begin{definition}[\bf Sample Mean] \label{Def_meanshift}
        Let $\mathcal{Q} \subset \mathcal{X}$ be a finite set of sample vectors in $\mathcal{X}$. Let $k:\mathbb{R}^+\rightarrow \mathbb{R}^+$ be a profile and $w: \mathcal{Q}\rightarrow \mathbb{R}^+$ be a weight function. 
        The sample mean with a profile $k$ at $x\in \mathcal{X}$ is defined by
        \begin{align*} 
            \varpi(x)=\frac{\sum_{q\in\mathcal{Q}}k\left(\|q-x\|^2\right)w\left(q\right)q}{\sum_{q\in\mathcal{Q}}k\left(\|q-x\|^2\right)w\left(q\right)}. 
        \end{align*}
    \end{definition}
\end{tcolorbox}

Consider an $n$-dimensional Euclidean space $\mathcal{X}$, and suppose that $\mathcal{Q}$ is a finite set of sample vectors in $\mathcal{X}$. 
The mean shift is defined by 
\begin{align*} 
    m(x)=\varpi(x)-x=\frac{\sum_{q\in\mathcal{Q}}k\left(\|q-x\|^2\right)w\left(q\right)q}{\sum_{q\in\mathcal{Q}}k\left(\|q-x\|^2\right)w\left(q\right)} -x
\end{align*}
where the definitions of profile $k(\cdot)$, weight $w(\cdot)$, and sample mean $\varpi(\cdot)$ are given in ``Key Definitions in Mean-Shift Theory''. 
Let $\mathcal{P}\subset \mathcal{X}$ denote a finite set of cluster centers. 
The evolution of $\mathcal{P}$ in the form of iterations $\mathcal{P}\leftarrow \mathcal{P}+m(\mathcal{P})$ with $m(\mathcal{P})=\{m(p):p\in\mathcal{P}\}$ is the so-called mean-shift procedure. 
The procedure halts if it finds a fixed point $\mathcal{P}=\mathcal{P}+m(\mathcal{P})$, i.e., $m(\mathcal{P})=0$. 

According to the above definitions, $\mathcal{P}$ and $\mathcal{Q}$ can be generalized to two separate sets with $\mathcal{Q}$ fixed through the process. 
In practice, one can use $\mathcal{P}$ to denote a set of robots, and $\mathcal{Q}$ to describe a set of feature points of the desired shape to which robots are expected to move. 
The set $\mathcal{Q}$  is fundamental to the design of collaboration strategies. 
Different definitions of $\mathcal{Q}$ can achieve different collaborative effects, such as precise formation and coverage formation. 
More details will be presented later. 

\subsection*{\emph{Properties of Mean Shift}}

The mean shift has three key properties. 

\subsubsection*{\emph{Parallelism}}

The first is parallelism \cite[Remark~1]{Cheng1995PAMI}, namely, all $p$ in $\mathcal{P}$ are simultaneously updated according to the previous $p$ and $\varpi(p)$ values. 
This property is crucial for practical applications of mean shift in swarm collaboration, which determines whether the mean shift can be deployed into a distributed swarm system. 
It is known that distributed systems are robust in practice and can support large-scale swarms. 

\subsubsection*{\emph{Smoothness}}

The second is smoothness \cite[Theorem~2]{Comaniciu2002PAMI}, namely the path of each $p$ in $\mathcal{P}$ follows a smooth trajectory, where the angle between two consecutive mean-shift vectors is always less than $\pi/2$. 
This property determines the motion efficiency, specifically to ensure a smooth motion of the swarm robots in shape-forming and switching tasks \cite{Sun2023NC}. 
Smooth motion can improve task efficiency and reduce energy consumption. 

\subsubsection*{\emph{Gradient Direction}}

The third is the gradient-direction property \cite[Theorem~1]{Comaniciu2002PAMI}, that is, the mean shift $m(x)$ always points to the direction where the sample-point density increases the most, as shown in Figure~\ref{Fig_meanshift}. 
This property gives robots the ability of selfless exploration: When a robot is surrounded by neighboring robots with nearby unoccupied locations, it would selflessly give up its current location by exploring the highest density of the unoccupied locations in the desired shape. 

Next, these properties are applied to achieve formation collaboration of robot swarms. 

\section*{Precise Shape Formation}

This section introduces the strategy of precise shape formation. 
Suppose the desired shape is specified by a set of goal locations $\mathcal{Q}=\{q_1,\ldots,q_{n_{\rm goal}}\}$, where $q_i \in \mathbb{R}^2$ and $\mathcal{Q}_i=\{q_k\in\mathcal{Q}: \|q_k-p_i\|<r_{\rm sense}\}$ denotes the local goal locations near robot $i$. 
The control objective is to govern the positions of robots $\{p_i\}_{i\in\mathcal{V}}$ such that $p_i=q_k$ for all $i\in\mathcal{V}$ and $p_i\neq p_j$ for all $i\neq j$. 
To simplify the problem, it is usually assumed that $n_{\rm goal}= n_{\rm robot}$. 

\subsection*{\emph{Goal Location Exploration}}

The process of assignment-free precise formation is highly dynamic because the goal location of each robot is indeterminate due to the absence of an explicit procedure of assignment and, more importantly, the avoidance of inter-robot competition for goal locations. 
If a confused robot can selflessly give up its current location and instead actively explore the highest density of unoccupied goal locations, then the goal competition can be avoided, thus achieving the formation. 

The following goal-exploring command, proposed in \cite{Zhang2024RAL}, can be used to achieve the above control objective by employing the adapted mean-shift algorithm: 
\begin{align} 
    v_i^{\rm exp}=\kappa_1 \frac{\sum_{q_k\in\mathcal{Q}_i}\varphi\left(\rho_k\right)\left(q_k-p_i\right)}{\sum_{q_k\in\mathcal{Q}_i}\varphi\left(\rho_k\right)} 
    \label{Equ_explore_pre}
\end{align}
where $\kappa_1>0$ is a constant and $\rho_k$ is the distribution density of robots at goal location $q_k$. 
The calculation of $\rho_k$ will be given later. 
The weight $\varphi(\cdot)$ can be found in \cite[Equation (15)]{Zhang2024RAL}. 
Here, $\varphi(\cdot)$ is monotonically decreasing from $1$ to $0$. 
The weight $\varphi(\rho_k)$ is large when $\rho_k$ is small, i.e., few robots around $q_k$, thus more weights are given to the goal locations with fewer surrounding robots. 

For command \eqref{Equ_explore_pre}, it is required to calculate the distribution density $\rho_k$ of robots at goal location $q_k$ (Algorithm~\ref{Alg_precise}, Line~12). The determination of $\rho_k$ requires considering two aspects (Figure~\ref{Fig_overview}(a)), namely, the distribution of robot $i$'s neighbors at $q_k$ and the impact of robot $i$ on $q_k$. 
Typically, the distribution density of robots at $q_k$ is defined by \cite{Zhang2024RAL}
\begin{align}
    \rho_k(q_k)=c_1\sum_{j\in\mathcal{N}_i}e^{-\|q_k-p_j\|^2/2\sigma^2} + \frac{c_2}{\pi}{\arccos\left(\cos\langle \nabla h_i, q_k-p_i \rangle\right)}, \quad p_i\neq q_k~{\rm or}~\nabla h_i\neq 0
    \label{Equ_density}
\end{align} 
where $c_1, c_2>0$ are constants, and $\cos\langle \cdot \rangle$ is a cosine operator. 
The settings of $c_1$ and $c_2$ follow the principle of $c_1$ first and then $c_2$. 
Their scales reflect the strength of two terms in \eqref{Equ_density}. 
The parameter $\sigma>0$ determines the sensitivity of $\rho_k$ to the distance between $p_j$ and $q_k$. 
If $p_i = q_k$, the second term in \eqref{Equ_density} is defined as $c_2/2$. 
If $\nabla h_i = 0$, only the first term in \eqref{Equ_density} is retained. 
The definition of hop-count gradient $\nabla h_i$ can be found in \cite[Equation (8)]{Zhang2024RAL}. 
In hop-count propagation, robots have two roles. 
The first is an anchoring robot, while all the goals around it are occupied by other robots. 
The anchoring robot plays a stubborn role by insisting on propagating zero in the swarm (Algorithm~\ref{Alg_precise}, Lines~5-7). 
Each of the rest non-anchoring robots updates its hop count by $h_i=\min_{j\in\mathcal{N}_i} h_j +1$ (Algorithm~\ref{Alg_precise}, Lines~8-11). 

The velocity command in \eqref{Equ_explore_pre} encourages robots to explore the unoccupied goal locations, as seen from the first term in \eqref{Equ_density}. 
A goal location unoccupied by robots means that the distribution density of robots around it is approximately zero. 
This way, the largest weight significantly contributes to the mean-shift command \eqref{Equ_explore_pre}. 
As the distribution density increases, the goal location becomes less important to robots because the first term in \eqref{Equ_density} increases. 
If the goal locations near a robot are all occupied, then the robot is blocked and becomes an anchor to propagate this conflict to other robots with zero (Algorithm~\ref{Alg_precise}, Lines~5-7). 
This information affects the density calculation of other unblocked robots, as seen from the second term in \eqref{Equ_density}. 
Specifically, an unblocked robot increases the distribution density of goal locations in the direction of the hop-count gradient, and the algorithm gives more weights to those in the opposite direction of the gradient (Algorithm~\ref{Alg_precise}, Lines~8-11). 
As a result, command \eqref{Equ_explore_pre} drives unblocked robots to explore the goal locations away from the blocked robots so as to provide more unoccupied goal locations for blocked robots. 

\begin{algorithm}[!t] 
    \caption{Precise Shape Formation.}
    \label{Alg_precise}
	\KwIn{$\mathcal{Q}=\{q_1,\ldots,q_{n_{\rm goal}}\}$, $v_{\rm max}$, $\sigma$, $c_1$, $c_2$, $\kappa_1$, $\kappa_2$, $\kappa_3$} 
    $h_i\leftarrow \infty$ \\
	\While{true} 
	{
		$msg\_{rec} \leftarrow$ all the messages received \\
		$\{p_j, h_j\} \leftarrow$ message in $msg\_{rec}$ \\ 
        \If{robot $i$ is blocked}
		{
            $h_i\leftarrow 0$ \\			
		}
		\Else
		{
            $h_i \leftarrow \min_{j\in\mathcal{N}_i} h_j +1$ \\
            $\nabla h_i \leftarrow$ hop-count gradient calculated by \cite[Equation (8)]{Zhang2024RAL} \\
        }        
        $\rho_k \leftarrow$ distribution density of robots calculated by \eqref{Equ_density} \\
        $q_{{\rm g},i}\leftarrow\arg\min_{q_k\in\mathcal{Q}} \|q_k-p_i\|$ \\
        $v_i\leftarrow$ velocity command calculated by \eqref{Equ_command_pre} \\
        $v_i\leftarrow$ limited by a threshold $v_{\rm max}$ \\
		robot $i$ moves with velocity command $v_i$ \\
		$msg\_{tra} \leftarrow\{p_i, h_i\}$ \\
		transmit $msg\_{tra}$
	}
\end{algorithm}

\subsection*{\emph{Precise Formation Control}}

To achieve precise shape formation, robots still need to handle two issues, namely how to interact with their neighbors and what to do when no goal locations are around them. 
The following integrated control law, proposed in \cite{Zhang2024RAL}, can be used to address these problems: 
\begin{align} 
    v_i=v_i^{\rm exp}+v_i^{\rm gui}+v_i^{\rm int}
    \label{Equ_command_pre}
\end{align}
where $v_i^{\rm gui}$ and $v_i^{\rm int}$ represent the  goal-guiding and neighbor-interacting velocity commands. 
Note that the calculation of $v_i$ only requires the neighbor's position $p_j$ and hop count $h_j$ (Algorithm~\ref{Alg_precise}, Lines~3-4). 
The mathematical definitions of $v_i^{\rm gui}$ and $v_i^{\rm int}$ are given as follows. 

If there are no goals around a robot, namely the robot is far away from the desired shape, it must select a goal location from $\mathcal{Q}$ and move toward it. 
For this purpose, the following goal-guiding command $v_i^{\rm gui}$, proposed in \cite{Zhang2024RAL}, is designed: 
\begin{align*} 
    v_i^{\rm gui}=\kappa_2 \frac{q_{{\rm g},i}-p_i}{\|q_{{\rm g},i}-p_i\|}
\end{align*}
where $\kappa_2$ is chosen from the set $\{0,v_{\rm con}\}$ and $v_{\rm con}$ is a positive constant. 
Specifically, if there are no goal locations around robot $i$, then $\kappa_2=v_{\rm con}$ and robot $i$ moves toward $q_{{\rm g},i}$ at a constant speed $v_{\rm con}$. 
By contrast, if $|\mathcal{Q}_i|\neq 0$, then $\kappa_2=0$. 
In this case, robot $i$ is free of $v_i^{\rm gui}$ and instead explores the unoccupied goals by the effect of $v_i^{\rm exp}$. 
The goal location $q_{{\rm g},i}$ for robot $i$ to move toward is defined as $q_{{\rm g},i}=\arg\min_{q_k\in\mathcal{Q}} \|q_k-p_i\|$, which means that robot $i$ selects the nearest location from $\mathcal{Q}$ as the goal point (Algorithm~\ref{Alg_precise}, Line~13). 

Whether there are goal locations around or not, the robot needs to avoid collisions with its neighbors. 
This objective can be achieved by the neighbor-interacting velocity command \cite{Zhang2024RAL} 
\begin{align} 
    v_i^{\rm int}=\kappa_3 \sum_{j\in\mathcal{N}_i} \mu \left(\|p_i-p_j\|\right)\left(p_i-p_j\right)
    \label{Equ_interact_pre}
\end{align}
where $\kappa_3$ is a positive constant. 
The definition of weight $\mu(\cdot)$ can be found in \cite[Equation (16)]{Sun2023NC}. 
By definition, $v_i^{\rm int}$ is a repulsion velocity that pushes robot $i$ away from its neighbors to avoid collisions. 
Specifically, if $\|p_i-p_j\|>0$, $\mu(\|p_i-p_j\|)=0$, and there is no avoidance interaction between $i$ and $j$. 
If $\|p_i-p_j\|<0$ and $\|p_i-p_j\|$ decreases, the weight $\mu(\|p_i-p_j\|)$ monotonically increases. 
The settings of $\kappa_1$, $\kappa_2$, and $\kappa_3$ follow the order of $\kappa_2$ first, then $\kappa_1$, and finally $\kappa_3$. 
The scale of a parameter reflects the strength of the corresponding behavior. 
For example, if $\kappa_1$ is larger, the robot will move toward its goal faster, and otherwise, the robot will move more slowly. More details are found in \cite{Zhang2024RAL}. 

\begin{figure}[!t]
    \centering
    \includegraphics[width=\linewidth]{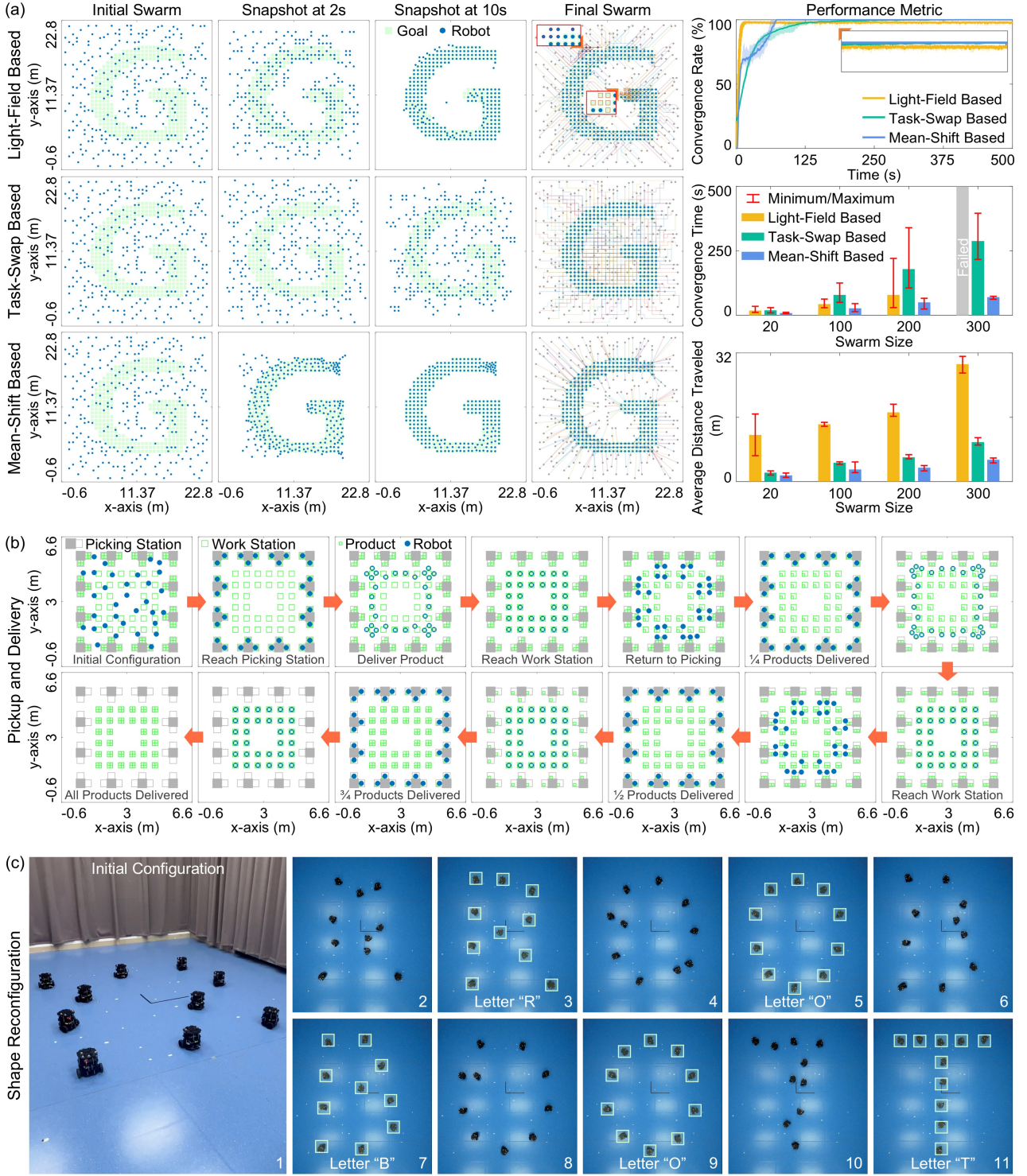}
    \caption{Results of precise formation. 
    (a)~Comparison between the mean-shift method \cite{Zhang2024RAL} and assignment-based methods \cite{Chu2023TASE,Wang2020TRO}. There are 300 robots forming a shape ``G''. The definitions of convergence rate and convergence time refer to \cite{Zhang2024RAL}.
    Another metric is the average distance traveled by all the robots from the initial moment to the convergence moment. 
    (b)~Simulation of 32 robots transferring products from picking stations to work stations. The parameters of (a) and (b) are listed as: $r_{\rm sense}=1.3$, $v_{\rm max}=2$, $c_1=1$, $c_2=0.4$, $\sigma=0.22$, $\kappa_1=3$, $\kappa_2=10$, and $\kappa_3=1$. 
    (c)~Experiment results of 10 real robots forming different shapes in a sequence. The parameters can be found in \cite{Zhang2024RAL}.} 
    \label{Fig_precise}
\end{figure}

Under the action of \eqref{Equ_command_pre}, robots are able to achieve a precise formation. 
A comparison with two state-of-the-arts \cite{Chu2023TASE,Wang2020TRO} is presented in Figure~\ref{Fig_precise}(a). 
It is shown that three methods can precisely assemble the user-specified shape from the same configurations when $n_{\rm robot} \leq 200$. As $n_{\rm robot} \leq 200$ increases to 300, the method in \cite{Chu2023TASE} easily falls into local minima known as \emph{deadlock}, as marked by the red box in Figure~\ref{Fig_precise}(a). 
In contrast to method \cite{Wang2020TRO}, the convergence time and average distance traveled of the mean-shift strategy in \cite{Zhang2024RAL} are both the smallest, and the convergence time is improved by at least 4 times. 
For demonstration, a group of 10 robots starting from a random configuration can form different shapes and switch from one to another smoothly, as shown in Figure~\ref{Fig_precise}(c). 

The mean-shift strategy finds important applications in smart warehousing, such as product pickup and delivery. 
In this application, robots are required to transfer products from specified picking stations to work stations (Figure~\ref{Fig_precise}(b)). This process can be visualized as robots achieving a shape specified by picking stations, picking up products, and then forming a shape specified by work stations. 
As shown in Figure~\ref{Fig_precise}(b), a group of 32 robots picks up products from picking stations and transfers them to work stations in a continuous cycle until all the products are transferred. 

\section*{Area Coverage Formation}

This section introduces the strategy of area coverage formation. In contrast to precise formation, coverage formation emphasizes a complete coverage of a desired shape area, but not a precise goal location in the shape. 
Suppose that the desired shape is specified by a binary grid. 
Each cell is described by $\varrho$ and $\xi_\varrho$. Here, $\varrho=(\varrho_x,\varrho_y)$ denotes the column and row indexes of the cell, and $\xi_\varrho\in[0,1]$ is the color of the cell: $\xi_\varrho=0$ if the cell is black and $\xi_\varrho=1$ if the cell is white. 
The black cells correspond to the desired shape that should be covered by the robots, i.e., $\mathcal{B}=\{\varrho:\xi_\varrho=0\}$. 

\subsection*{\emph{Shape Area Exploration}}

The control objective is to govern the positions of robots such that the area occupied by all the robots covers the desired shape completely. 
This objective can be achieved by an adaptive version of the mean-shift algorithm \cite{Sun2023NC}. 
The key idea is that, when a robot is surrounded by neighbors and unoccupied areas, it will selflessly give up its current occupation and explore the highest density of unoccupied area in the shape. 
For that purpose, the following shape-exploring velocity command $v_i^{\rm exp}$ is defined \cite{Sun2023NC}: 
\begin{align} 
    v_i^{\rm exp}=\kappa_1 \frac{\sum_{\varrho\in\mathcal{C}_i}\varphi\left(\|q_\varrho-p_i\|/r_{\rm sense}\right)\left(q_\varrho-p_i\right)}{\sum_{\varrho\in\mathcal{C}_i}\varphi\left(\|q_\varrho-p_i\|/r_{\rm sense}\right)} 
    \label{Equ_explore_cov}
\end{align}
where $\kappa_1>0$ is a constant and $q_\varrho$ is the position of cell $\varrho$. 
Here, $\varrho$ belongs to the set $\mathcal{C}_i$, and $\mathcal{C}_i$ denotes the set of black cells around robot $i$. 
The details about $\mathcal{C}_i$ will be presented later. 
The function $\varphi(\cdot)$ is defined in the same way as \eqref{Equ_explore_pre}. 
Generally, more weights are given to the cells that are closer to robot $i$. 

For command \eqref{Equ_explore_cov}, it is required to calculate the cell position $q_\varrho$, namely how to transform a cell index $\varrho$ into the global reference frame (Algorithm~\ref{Alg_coverage}, Lines~4-6). 
Let $\varrho_o$ be the cell located closest to the center of the binary shape, and $q_{\varrho_o}$ be the position of $\varrho_o$ in the global frame (Algorithm~\ref{Alg_coverage}, Line~1). 
Then, the cell position $q_\varrho$ is given by $q_\varrho=(\varrho-\varrho_o)\ell_{\rm cell}+q_{\varrho_o}$. 
Here, $q_{\varrho_o}$ is essential, but the shape orientation is ignored. 

Next, the objective is to parameterize the cell size $\ell_{\rm cell}$ (Algorithm~\ref{Alg_coverage}, Line~2). 
Let $n_{\rm cell}$ be the number of black cells in the binary shape. 
On the one hand, the area of all the black cells is $n_{\rm cell}\ell_{\rm cell}^2$. 
On the other hand, the total area occupied by the robots is $\pi(r_{\rm avoid}/2)n_{\rm robot}$. 
Here, the space occupied by any robot $i$ is approximated by a circle with the center at $p_i$ and the radius of $r_{\rm avoid}/2$. 
In coverage formation, these two areas are expected to be equal such that the robots are able to cover the desired shape, i.e., $n_{\rm robot} \pi(r_{\rm avoid}/2)^2 =n_{\rm cell}\ell_{\rm cell}^2$. 
As a result,  the cell size is $\ell_{\rm cell}=\sqrt{\frac{\pi}{4}\frac{n_{\rm robot}}{n_{\rm cell}}}r_{\rm avoid}$. 

The velocity command in \eqref{Equ_explore_cov} plays two vital roles in shape formation processes. 
First, if robot $i$ is inside the shape, the command \eqref{Equ_explore_cov} encourages robot $i$ to explore the unoccupied areas of the shape (Figure~\ref{Fig_overview}(b)). 
In this case, $\mathcal{C}_i$ is defined as the set of all the unoccupied black cells that are within the sensing radius $r_{\rm sense}$ (Algorithm~\ref{Alg_coverage}, Lines~9-10). 
At this time, $\kappa_1$ is taken as $\sigma_1$. 
Second, if robot $i$ is close to the boundary of the desired shape, the command \eqref{Equ_explore_cov} pulls robot $i$ into the shape. At this time, $\kappa_1$ is taken as $\sigma_2$. 
Here, $\mathcal{C}_i$ is defined as the set of all the black cells that are within the sensing radius $r_{\rm sense}$, whether they are occupied or not (Algorithm~\ref{Alg_coverage}, Lines~13-14). 

\begin{algorithm}[!t]
    \caption{Area Coverage Formation.}
    \label{Alg_coverage}
	\KwIn{$\mathcal{B}=\{\varrho:\xi_\varrho=0\}$, $v_{\rm max}$, $q_{\varrho_o}$, $\sigma_1$, $\sigma_2$, $\kappa_2$, $\kappa_3$} 
    $\varrho_o \leftarrow$ cell located closest the center of $\mathcal{B}$ \\
    $\ell_{\rm cell}\leftarrow\sqrt{(\pi n_{\rm robot}) / (4 n_{\rm cell})} r_{\rm avoid}$ \\
    $\mathcal{B}_{\rm gray}\leftarrow$ gray conversion of $\mathcal{B}$ by \cite[Equation~(1)]{Sun2023NC}\\    
    \For{$\varrho$ in $\mathcal{B}_{\rm gray}$}
    {
        $q_\varrho\leftarrow(\varrho-\varrho_o)\ell_{\rm cell}+q_{\varrho_o}$
    }
	\While{true} 
	{		
        $\varrho_i \leftarrow$ index of the cell that is closest to $p_i$ \\
        \If{$\xi_{\varrho_i} == 0$}
		{
            $\mathcal{C}_i\leftarrow$ set of all the unoccupied black cells around robot $i$ \\
            $\varrho_{{\rm g},i}\leftarrow \varrho_i$ \\
		}
		\Else
		{
            $\mathcal{C}_i\leftarrow$ set of all the black cells around robot $i$ \\
            \If{$\xi_{\varrho_i} == 1$}
            {
                $\varrho_{{\rm g},i}\leftarrow$ the nearest gray cell around $\varrho_i$
            }
            \Else
            {
                $\varrho_{{\rm g},i}\leftarrow$ the darkest cell in the $3\times 3$ mask centered at $\varrho_i$
            }           
        } 
        $q_{{\rm g},i}\leftarrow$ position of cell $\varrho_{{\rm g},i}$ \\        
        $msg\_{rec} \leftarrow$ all the messages received \\
		$\{p_j, v_j\} \leftarrow$ message in $msg\_{rec}$ \\ 
        $v_i\leftarrow$ velocity command calculated by \eqref{Equ_command_cov} \\
        $v_i\leftarrow$ limited by a threshold $v_{\rm max}$ \\
		robot $i$ moves with velocity command $v_i$ \\
		$msg\_{tra} \leftarrow\{p_i, v_i\}$ \\
		transmit $msg\_{tra}$
	}
\end{algorithm}

\subsection*{\emph{Coverage Formation Control}}

The following control law, proposed in \cite{Sun2023NC}, can be used to solve the coverage formation problem: 
\begin{align}
    v_i=v_i^{\rm exp}+v_i^{\rm acc}+v_i^{\rm int}
    \label{Equ_command_cov}
\end{align}
where $v_i^{\rm acc}$ and $v_i^{\rm int}$ denote the shape-accessing and neighbor-interacting velocity commands, respectively. 
Note that the calculation of $v_i$ only requires the neighbor's position $p_j$ and velocity $v_j$ (Algorithm~\ref{Alg_coverage}, Lines~23-24). 
The definitions of $v_i^{\rm acc}$ and $v_i^{\rm int}$ are given as follows. 

To achieve area coverage formation, robots also need to determine what to do when they are far away from the shape. 
In this case, robots have no black cells around but need to move toward the desired shape as quickly as possible, which can be achieved by the shape-accessing command \cite{Sun2023NC} 
\begin{align*} 
    v_i^{\rm acc}=\kappa_2 \xi_{\varrho_i} \frac{q_{{\rm g},i}-p_i}{\|q_{{\rm g},i}-p_i\|}
\end{align*}
where $\kappa_2>0$ is a constant and $q_{{\rm g},i}$ is the local goal location for robot $i$ to move toward. 
Recall that $\xi_{\varrho_i}$ is the color of cell $\varrho_i$. 
Here, $\varrho_i$ is the index of the cell that is closest to $p_i$. 
The calculations of $\xi_{\varrho_i}$ and $q_{{\rm g},i}$ can be found in \cite[Section~3 in Supplementary Information]{Sun2023NC}. 
The goal location $q_{{\rm g},i}$ is the position of cell $\varrho_{{\rm g},i}$. 
The local goal cell $\varrho_{{\rm g},i}$ is the darkest cell around robot $i$. 
More specifically, if the cell $\rho_i$ closest to robot $i$ is inside the grayscale shape, $\varrho_{{\rm g},i}$ is the darkest cell in the $3\times 3$ mask centered at robot $i$ (Algorithm~\ref{Alg_coverage}, Lines~18-20). 
If $\rho_i$ is outside the grayscale shape, $\varrho_{{\rm g},i}$ is the nearest gray cell around robot $i$ (Algorithm~\ref{Alg_coverage}, Lines~15-17). 
In summary, $v_i^{\rm acc}$ aims to drive robots to access the desired shape along the gray gradient, which is necessary when a robot has no black cells within its radius $r_{\rm sense}$, and it fills the role of being unable to explore the desired shape by using the shape-exploring command when robots are far away from the shape. 

The task of the neighbor-interacting command in \eqref{Equ_command_cov} is to achieve collision avoidance and velocity alignment with neighboring robots. 
Define \cite{Sun2023NC} 
\begin{align*} 
    v_i^{\rm int}=\kappa_3 \sum_{j\in\mathcal{N}_i} \mu \left(\|p_i-p_j\|\right)\left(p_i-p_j\right)-\sum_{j\in\mathcal{N}_i} \frac{1}{|\mathcal{N}_i|} \left(v_i-v_j\right)
\end{align*}
where $\kappa_3$ is a positive constant and $\mu$ is a weight function defined as \eqref{Equ_interact_pre}. 
The first term is a repulsion command that pushes robot $i$ away from their neighbors to avoid collisions. 
The second term is a consensus command that aims to align robot $i$'s velocity with its neighbors. 
The settings of $\kappa_1$, $\kappa_2$, and $\kappa_3$ also follow the order of $\kappa_2$ first, then $\kappa_1$, and finally $\kappa_3$, the scales of which reflect the strength of the corresponding behavior. More details are found in \cite{Sun2023NC}. 

\begin{figure}[!t]
    \centering
    \includegraphics[width=\linewidth]{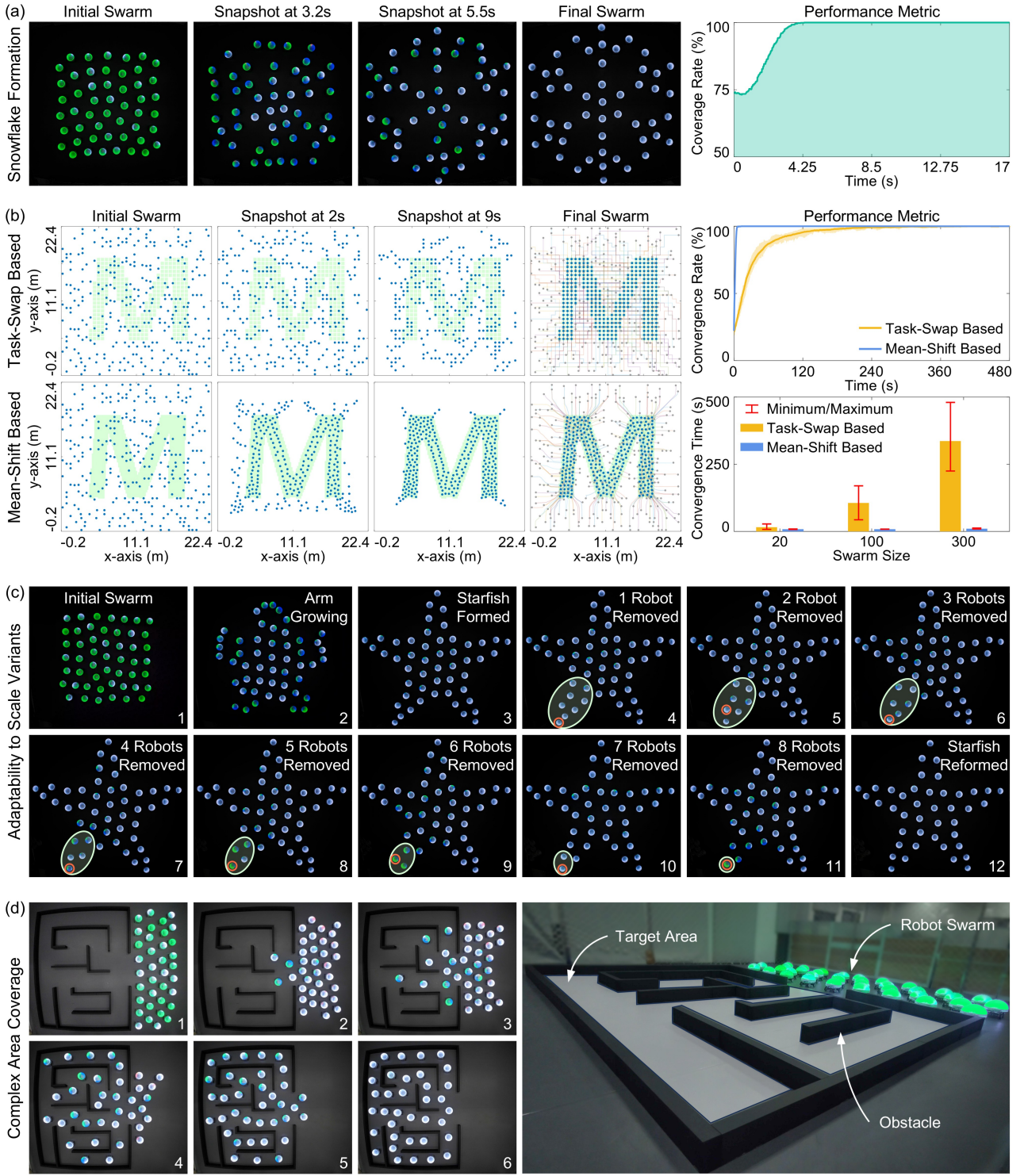}
    \caption{Results of coverage formation. 
    (a)~Experiment results of 50 robots forming a snowflake shape. The definition of coverage rate refers to \cite{Sun2023NC}. 
    (b)~Comparison between the mean-shift method \cite{Sun2023NC} and the assignment-based method \cite{Wang2020TRO}. There are 300 robots forming a shape ``M''. The definitions of convergence rate and convergence time refer to \cite{Sun2023NC}. 
    (c)~Experiment results of 50 robots forming a starfish shape. Even as robots are sequentially removed from the shape, the rest of the robots can form a new starfish shape. 
    (d)~Experiment results of 36 robots exploring a complex maze without getting deadlocks at any corners. 
    The simulation parameters of (b) are listed as: $r_{\rm sense}=1.3$, $v_{\rm max}=2$, $q_{\varrho_o}=[11.1, 11.1]^{\rm T}$, $\sigma_1=2$, $\sigma_2=10$, $\kappa_2=4$, and $\kappa_3=20$. 
    The experiment parameters of (a), (c), and (d) can be found in \cite{Sun2023NC}.} 
    \label{Fig_coverage}
\end{figure}

The control strategy \eqref{Equ_command_cov} is able to achieve the coverage formation for complex shapes. 
One representative shape is the snowflake shape (Figure~\ref{Fig_coverage}(a)). 
Moreover, the mean-shift strategy \cite{Sun2023NC} can greatly improve the efficiency compared with the state-of-the-art method \cite{Wang2020TRO}, and this improvement is more obvious as the swarm scale increases (Figure~\ref{Fig_coverage}(b)). 
Specifically, as the swarm size increases to 300, the convergence time of mean-shift strategy proposed \cite{Sun2023NC} is at least 20 times faster than method \cite{Wang2020TRO}. 
This is primarily because mean shift exploration does not require goal assignment, whereas the method in \cite{Wang2020TRO} needs to execute local goal swaps constantly. 
In addition to efficiency, the mean-shift strategy in \cite{Sun2023NC} also exhibits strong adaptability to individual faults (Figure~\ref{Fig_coverage}(c)). 

The mean-shift strategy has promising potential in complex area exploration. 
In this application, robots are required to evenly fulfill an unknown area while avoiding obstacles, which can be mapped to area coverage formations. 
As a representative example, a group of 36 robots can explore a complex maze (Figure~\ref{Fig_coverage}(d)). 
Although there are many obstacles in the maze, the robot swarm can explore and cover the maze successfully without getting deadlocks at any corners. 

\section*{Maneuvering Formation}

Previously, the shape position and orientation are subjectively specified. 
However, the algorithms are difficult to cope with maneuvering formations since the shape position and orientation are time-varying. 
To address this issue, the strategy proposed in \cite{Sun2023NC} is to let the robots negotiate them and reach consensus in a distributed manner. 

Let $q_{o,i}$ and $\phi_{o,i}$ denote the shape position and orientation of robot $i$, respectively. 
The following negotiation protocol can drive robots to reach a consensus \cite{Sun2023NC}: 
\begin{align} 
    \nu_{o,i}=-c_1 \sum_{j\in \mathcal{N}_i} \frac{1}{|\mathcal{N}_i|} {\rm sign}\left(q_{o,i}-q_{o,j}\right) \left|q_{o,i}-q_{o,j}\right|^{\alpha} + \sum_{j\in\mathcal{N}_i} \frac{1}{|\mathcal{N}_i|} \nu_{o,j}
    \label{Equ_uninformed_pos}
\end{align}
where $c_1>0$ and $0<\alpha<1$ are two constants. 
Here, ${\rm sign}(\cdot)$ and $|\cdot|$ represent the component-wise sign and absolute value of a real vector, respectively. 
There are two terms in \eqref{Equ_uninformed_pos}. 
The first term is the deficiency of the position between robot $i$ and its neighbors, which aims to drive $q_{o,i}\rightarrow q_{o,j}$. 
The second term is the average velocity of the neighbors, aiming to drive $\nu_{o,i}\rightarrow \nu_{o,j}$. 
Since $0<\alpha<1$, negotiations can be achieved in a finite time. 
Regarding the shape orientation, the following protocol can be used to perform negotiation \cite{Sun2023NC}: 
\begin{align} 
    \omega_{o,i}=-c_2 \sum_{j\in \mathcal{N}_i} \frac{1}{|\mathcal{N}_i|} {\rm sign}\left(\phi_{o,i}-\phi_{o,j}\right) \left|\phi_{o,i}-\phi_{o,j}\right|^{\alpha} + \sum_{j\in\mathcal{N}_i} \frac{1}{|\mathcal{N}_i|} \omega_{o,j}
    \label{Equ_uninformed_ori}
\end{align}
where $c_2>0$ is a constant. 
As can be seen, protocol \eqref{Equ_uninformed_ori} has the same structure as \eqref{Equ_uninformed_pos}, where the first term in \eqref{Equ_uninformed_ori} aims to drive $\phi_{o,i}\rightarrow \phi_{o,j}$ and the second term aims to drive $\omega_{o,i}\rightarrow \omega_{o,j}$. 

To handle maneuvering formations, it is necessary to introduce a small number of informed robots who know the trajectory reference of the shape. 
The informed robots play a stubborn role by insisting on their desired trajectories, while the others gradually converge to the informed ones. 
In particular, informed robots are driven by \cite{Sun2023NC} 
\begin{align} 
    \nu_{o,i}=-c_1 \sum_{j\in \mathcal{N}_i} \frac{1}{|\mathcal{N}_i|} {\rm sign}\left(q_{o,i}-q_{o,j}\right) \left|q_{o,i}-q_{o,j}\right|^{\alpha} + c_3 \left(q_{\rm ref}-q_{o,i}\right) + \nu_{\rm ref}
    \label{Equ_informed_pos}
\end{align}
where $c_3>0$ is a constant. 
Here, $q_{\rm ref}$ and $\nu_{\rm ref}$ are the position and velocity of the time-varying reference, respectively. 
The aim of the first term in \eqref{Equ_informed_pos} is the same as that in \eqref{Equ_uninformed_pos}. 
The second term in \eqref{Equ_informed_pos} is to track the reference trajectories. 
Regarding the shape orientation negotiation, informed robots are driven by \cite{Sun2023NC} 
\begin{align} 
    \omega_{o,i}=-c_2 \sum_{j\in \mathcal{N}_i} \frac{1}{|\mathcal{N}_i|} {\rm sign}\left(\phi_{o,i}-\phi_{o,j}\right) \left|\phi_{o,i}-\phi_{o,j}\right|^{\alpha} + c_4 \left(\phi_{\rm ref}-\phi_{o,i}\right) + \omega_{\rm ref}
    \label{Equ_informed_ori}
\end{align}
where $c_4>0$ is a constant and $\phi_{\rm ref},\omega_{\rm ref}$ are the orientation and angular speed of the time-varying reference. 
The protocol \eqref{Equ_informed_ori} has the same structure as \eqref{Equ_informed_pos}, the purpose of which is to drive $\phi_{o,i}\rightarrow \phi_{\rm ref}$ and $\omega_{o,i}\rightarrow \omega_{\rm ref}$. 
Note that $c_1$, $c_2$, $c_3$, and $c_4$ are insensitive because the negotiation is free of robot dynamics. 
In experience, they are equal. 
The convergence analysis of \eqref{Equ_uninformed_pos}-\eqref{Equ_informed_ori} is given in \cite[Theorem 1 in Supplementary Information]{Sun2023NC}. 

\begin{figure}[!t]
    \centering
    \includegraphics[width=\linewidth]{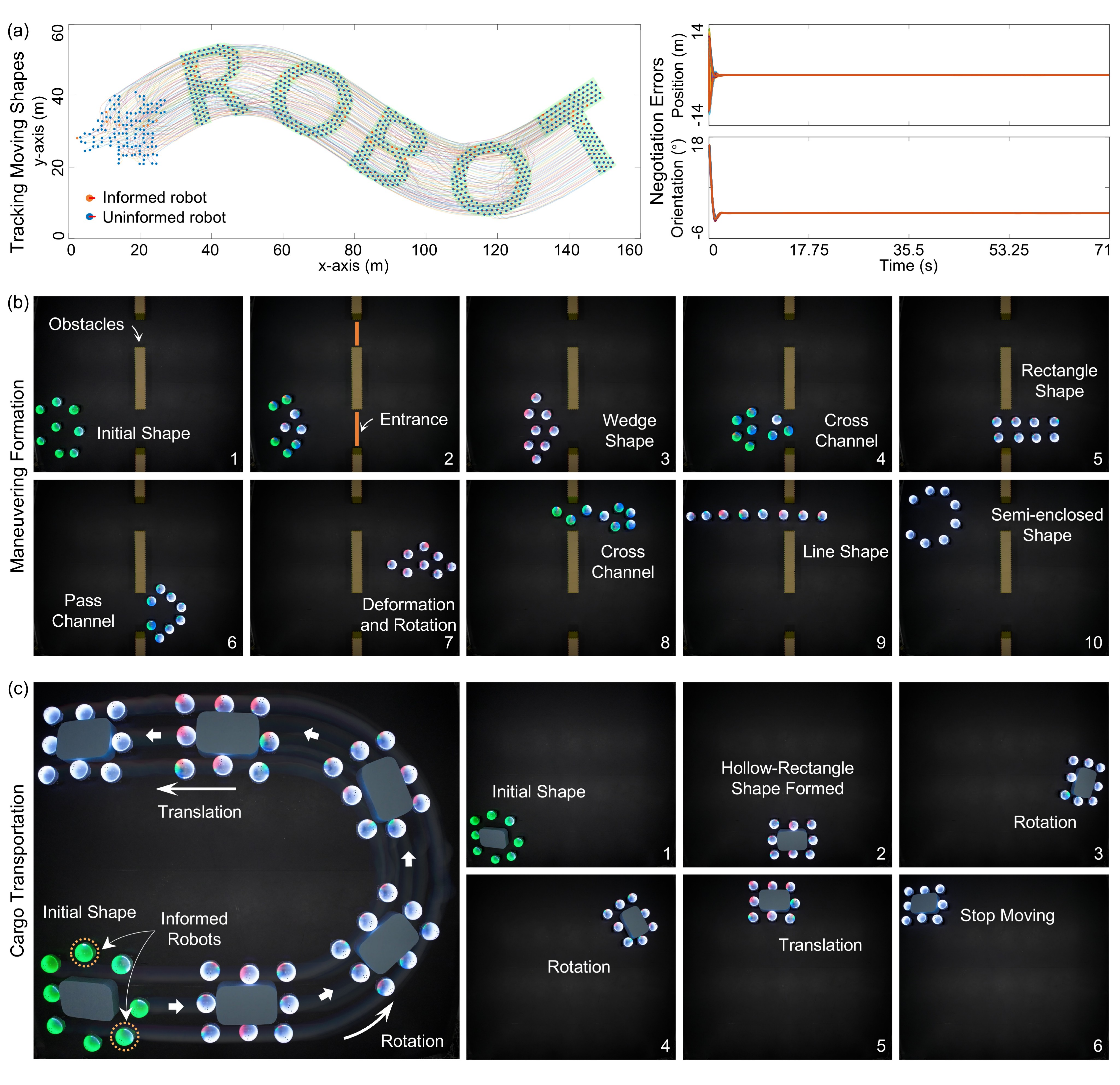}
    \caption{Results of maneuvering formation. 
    (a)~Simulation results of 128 robots tracking moving shapes in a sequence. 
    (b)~Experiment results of 8 robots forming different shapes, including wedge, rectangle, line, and semi-enclosed shapes, in a sequence. 
    (c)~Cargo transportation by a group of 8 real robots. 
    The simulation and experiment parameters are consistent with those in Figure~\ref{Fig_coverage}.
    The negotiation parameters are listed as: $c_1=c_2=c_3=c_4=1.6$, and $\sigma=0.8$.} 
    \label{Fig_maneuver}
\end{figure}

With protocols \eqref{Equ_uninformed_pos}-\eqref{Equ_informed_ori}, robots can move collectively while maintaining the desired shape, and reach a consensus on the time-varying trajectory (Figure~\ref{Fig_maneuver}(a)). 
For demonstration, a group of 8 robots can keep maneuvering to change their position, orientation, and geometric shape to avoid obstacles (Figure~\ref{Fig_maneuver}(b)). 
In addition, the mean-shift strategy is essential for robots to achieve cargo transportation. 
In this application, robots are required to surround and transport the cargo that is larger than an individual's size. 
As shown in Figure~\ref{Fig_maneuver}(c), a group of 8 robots can form a hollow moving shape to automatically transport the cargo while encircling it in the center position. 

\section*{Conclusion and Future Lookout}

This article has presented a review of the mean-shift theory and its applications in distributed shape formation. 
Motivated by the fact that most existing works adopt goal assignments that may cause critical limitations in efficiency and robustness, this article has demonstrated how to apply the mean-shift theory to solve the problem of shape formation in large swarms more effectively. 
Three specific problems have been discussed, namely precise formation, coverage formation, and maneuvering formation. 

The emerging research area of mean-shift based collaboration is far from being fully explored. 
One limitation is that the desired shape is assumed to be single-connected. 
This is because the mean-shift strategy relies on local sensing of cells. If two disconnected shapes are far apart, it becomes difficult for a robot to explore cells across different shapes. 
This assumption may not be valid in practice, and thus it is necessary to study the case of multi-shape formation. 
Another limitation is that either goal-location or binary-grid representation requires storing thousands or even tens of thousands of points, which imposes higher demands on onboard storage. 
Thus, it is important to develop low-dimensional representation methods that are independent of swarm size. 
Third, it is necessary to explore new dimensions of performance evaluation, especially in terms of energy consumption, communication overhead, and robustness. 
In addition, the following important problems remain open to be solved. 

\subsection*{\emph{Adaptive Shape Decision}}

One key feature of the results in this article is that the desired shape is subjectively specified, which means that all the robots must know the desired shape to be formed or deformed. 
This premise is often impractical because robots need to adaptively reconfigure their shape in response to obstacle-dense environments. 
Therefore, it is meaningful to enable the robots the ability of adaptive shape decisions. 
If the shape decision is taken into account, however, the problem will become more complicated because undesired equilibria may appear in decision-making based on local information. 
Despite the recent advances in shape decisions for centralized optimization \cite{Yang2022NMI}, the problem of distributed decision-making for large robot swarms is still a technical challenge to address. 

\subsection*{\emph{Autonomous Localization}}

Another necessary condition imposed by the methods discussed here is that an external localization system, such as GPS, is required to locate the robots. 
External localization systems, however, may not be accessible in many scenarios, such as indoors or inside building atriums. 
In contrast to relying on external localization systems, robots equipped with onboard sensors for autonomous localization are more reliable and autonomous. 
Thus, it is meaningful to study shape formation in the absence of external localization systems. 
One trivial approach is to allow robots to satisfy rigid conditions, such as distance keeping \cite{Shen2023TCNS} or bearing rigidity \cite{Li2022TC}, to achieve autonomous localization. 
This rigid condition, however, is difficult to satisfy in practice due to the limited sensor measurements and external disturbances. 
The problem of how to relieve rigid conditions remains open to be solved. 

\subsection*{\emph{Machine Learning for Robot Swarm}} 

Recently, machine learning has undergone great advancement with the increase in data availability and computing power \cite{Yang2024NMI}. 
In particular, machine learning is revolutionizing robotics by enhancing perception, decision-making, and control capabilities, enabling the robot to operate in complex scenarios beyond the traditional methods. 
However, the upsizing of swarm size to multiple and even large scale poses new challenges, such as emergent complexities in robotic interactions, strong nonlinearity in collective decision-making, dynamic uncertainty in the environment, and so on. 
As a consequence, the problem of how to integrate machine learning into multi-robot collaborative systems will emerge as a significant research trend. 

Mean-shift theory and its applications to shape formation are promising for large-scale collaborative robot swarms, therefore are expected to be further developed, especially in smart systems and industrial automation. 

\section*{Acknowledgements}

This work was supported in part by the National Natural Science Foundation of China under Grant 62503028, in part by the National Key Research and Development Program of China under Grant 2022YFB3305600, and in part by the National Key Laboratory
of Multi-Perch Vehicle Driving Systems under Grant QDXT-NZ-202407-01. 
The corresponding author is Jinhu L\"{u}. 

\section*{Biographies}

\textbf{\emph{Guibin Sun}} (sunguibinx@buaa.edu.cn) received his Ph.D. degree in control science and engineering from Beihang University, Beijing, China, in 2022. 
From 2022 to 2025, he was a postdoctoral research fellow of “Zhuoyue” program with Beihang University, Beijing, China. 
He is currently an associate professor with the School of Automation Science and Electrical Engineerin, Beihang University, Beijing, China. 
He has authored over 20 research papers in international journals, including Nature Communications and IEEE Transactions. 
His research interests include theories and applications of robot swarms, especially for large-scale swarms. 

\textbf{\emph{Jinhu L\"{u}}} (jhlu@iss.ac.cn) received his Ph.D. degree in applied mathematics from the Academy of Mathematics and Systems Science, Chinese Academy of Sciences, Beijing, China, in 2002. 
He was a professor with RMIT University, Melbourne, VIC, Australia. 
Currently, he is a vice-president for scientific research with Beihang University, Beijing, China. 
His research interests include cooperation control, industrial internet, complex networks, and big data. 
He was the general co-chair of the 43rd Annual Conference of the IEEE Industrial Electronics Society in 2017. He was a co-editor-in-chief of IEEE Transactions on Industrial Informatics. 
He is a Fellow of IEEE. 

\textbf{\emph{Kexin Liu}} (kxliu@buaa.edu.cn) received his Ph.D. degree in system theory from the Academy of Mathematics and Systems Science, Chinese Academy of Sciences, Beijing, China, in 2016. 
From 2016 to 2018, he was a postdoctoral fellow with Peking University, Beijing, China. 
From 2018 to 2024, he was an associate professor with Beihang University, Beijing, China. 
Currently, he is a professor with the School of Automation Science and Electrical Engineering, Beihang University, Beijing, China. 
His research focuses on multiagent systems and complex networks. 

\textbf{\emph{Zhenqian Wang}} (zhenqianwang@buaa.edu.cn) received his Ph.D. degree in electrical engineering from Clemson University, Clemson, SC, USA, in 2020. 
He is currently an associate professor with the School of Automation Science and Electrical Engineering at Beihang University. His research interests include distributed cooperative control of multi-agent systems, fault tolerant control, and privacy preserving of cyber-physical systems.

\textbf{\emph{Guanrong Chen}} (eegchen@cityu.edu.hk) received his Ph.D. degree in applied mathematics from Texas A\&M University, TX, USA, in 1987. 
Since 2000, he has been a chair professor and the founding director of the Centre for Complexity and Complex Networks at the City University of Hong Kong. He is currently the Hong Kong Shun Hing Education and Charity Fund Chair Professor in Engineering. His research interests are in the fields of complex networks, nonlinear systems dynamics and control, and has been a highly cited searcher in engineering continuously for some ten years. 
He is a Life Fellow of IEEE. 

\bibliography{references}
\bibliographystyle{IEEEtran}

\end{document}